\documentclass[11pt,a4paper]{article}

\usepackage[dvipsnames]{xcolor}
\usepackage[hyperref]{acl2019}

\usepackage{subcaption}
\usepackage{mwe}
\usepackage{booktabs} 
\usepackage{svrsymbols}

\usepackage{times}
\usepackage{latexsym}
\usepackage{graphicx} 
\usepackage{tikz-dependency}

\aclfinalcopy 

\title{Word Interdependence Exposes How LSTMs Compose Representations}

\author{Naomi Saphra \\
  University of Edinburgh \\
  \texttt{n.saphra@ed.ac.uk} \\\And
  Adam Lopez \\
  University of Edinburgh \\
  \texttt{alopez@inf.ed.ac.uk} \\}

\date{}

\begin{document}
\maketitle

\begin{abstract}
Recent work in NLP shows that LSTM language models capture compositional structure in language data. For a closer look at how these representations are composed hierarchically, we present a novel measure of interdependence between word meanings in an LSTM, based on their interactions at the internal gates. To explore how compositional representations arise over training, we conduct simple experiments on synthetic data, which illustrate our measure by showing how high interdependence can hurt generalization. These synthetic experiments also illustrate a specific hypothesis about how hierarchical structures are discovered over the course of training: that parent constituents rely on effective representations of their children, rather than on learning long-range relations independently. We further support this measure with experiments on English language data, where interdependence is higher for more closely syntactically linked word pairs.
\end{abstract}

\section{Introduction}

Recent work in NLP has seen a flurry of interest in the question: are the representations learned by neural networks compositional? That is, are representations of longer phrases built recursively from representations of shorter phrases, as they are in many linguistic theories? If so, how and when do they learn to do this?

Computationally, sequence models like LSTMs scan a sentence from left to right, accumulating meaning into a hidden representation at each time step. Yet we have extensive evidence that fully trained LSTMs are sensitive to syntactic structure, suggesting that they learn something about recursive composition of meaning. For example, they can recall more history in natural language data than in similarly Zipfian-distributed n-gram data, implying that they exploit long-distance dependencies \citep{liu_lstms_2018}. Their internal representations seem to be hierarchical in nature \citep{blevins_deep_2018,hupkes_visualisation_2017}. They seemingly encode knowledge of part of speech \citep{belinkov_what_2017}, morphological productivity \citep{Vania2017FromCT}, and verb agreement \citep{lakretz_emergence_2019}. How does this apparently compositional behavior arise in learning?

Concretely, we are interested in an aspect of compositionality sometimes called \emph{localism}~\citep{hupkes_compositionality_2019}, in which meanings of long sequences are recursively composed from meanings of shorter child sequences, without regard for how the child meaning is itself constructed---that is, the computation of the composed meaning relies only on the local properties of the child meanings. By contrast, a global composition would be constructed from all words. A local composition operation leads to hierarchical structure (like a classic syntax tree), whereas a global operation leads to flat structure. If meaning is composed locally, then in a sentence like ``The chimney sweep has sick lungs'', the unknown composition function $f$ might reflect syntactic structure, computing the full meaning as: $f(f($The, chimney, sweep$)$, has, $f($sick, lungs$))$. Local composition assumes low interdependence between the meanings of ``chimney'' and ``has'', or indeed between any pair of words not local to the same invocation of $f$.

To analyze compositionality, we propose a measure of word interdependence that directly measures the composition of meaning in LSTMs through the interactions between words (Section~\ref{sec:methods}). Our method builds on Contextual Decomposition~\citep[CD; ][]{murdoch_beyond_2018}, a tool for analyzing the representations produced by LSTMs. We conduct experiments on a synthetic corpus (Section~\ref{sec:synthetic}), which illustrate interdependence and find that highly familiar constituents make nearby vocabulary statistically dependent on them, leaving them vulnerable to the domain shift. We then relate word interdependence in an English language corpus (Section~\ref{sec:english}) to syntax, finding that word pairs with close syntactic links have higher interdependence than more distantly linked words, even stratifying by sequential distance and part of speech. This pattern offers a potential structural probe that can be computed directly from an LSTM without learning additional parameters as required in other methods \citep{hewitt_structural_nodate}.

\section{Methods}\label{sec:methods}

We now introduce our interdependence measure, a natural extension of Contextual Decomposition~\citep[CD; ][]{murdoch_beyond_2018}, a tool for analyzing the representations produced by LSTMs. To conform with \citet{murdoch_beyond_2018}, all experiments use a one layer LSTM, with inputs taken from an embedding layer and outputs processed by a softmax layer.

\subsection{Contextual Decomposition} \label{sec:cd} 

Let us say that we need to determine when  our language model has learned that ``either'' implies  an appearance of ``or'' later in the sequence. We consider an example sentence, ``\textit{Either} Socrates is mortal \textit{or} not''.  Because many nonlinear functions are applied in the intervening span ``Socrates is mortal'', it is difficult to directly measure the influence of ``either'' on the later occurrence of ``or''. To dissect the sequence and understand the impact of individual elements in the sequence, we could employ CD.

\begin{figure}
    \centering
    \includegraphics[width=0.48\textwidth]{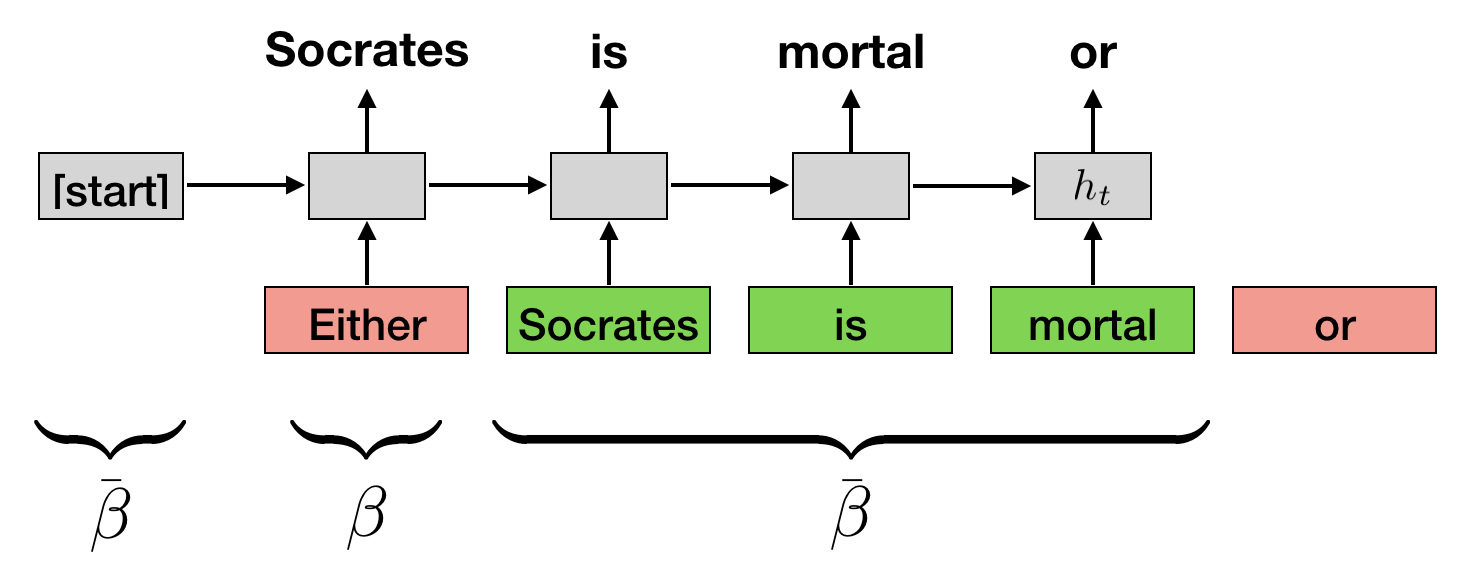}
    \caption{CD uses linear approximations of gate operations to linearize  the sequential application of the LSTM module.}
    \label{fig:cd_example}
\end{figure}

CD is a method of looking at the individual influences that words and phrases in a sequence have on the output of a recurrent model. Illustrated in Figure~\ref{fig:cd_example},  CD decomposes the activation vector produced by an LSTM layer into a sum of relevant and irrelevant parts. The \textbf{relevant} part is the contribution of the phrase or set of words \textbf{in focus}, i.e., a set of words whose impact we want to measure. We denote this contribution as $\beta$. The \textbf{irrelevant} part includes the contribution of all words not in that set (denoted $\bar{\beta}$) as well as interactions between the relevant and irrelevant words (denoted $\beta \interaction \bar{\beta}$). For an output hidden state vector $h$ at any particular timestep, CD will decompose it into two vectors: the relevant $h_{\beta}$, and irrelevant $h_{\bar{\beta}; \beta \interaction \bar{\beta}}$, such that:
$$h \approx h_{\beta} + h_{\bar{\beta}; \beta \interaction \bar{\beta}}$$
This decomposed form is achieved by linearizing the contribution of the words in focus. This is necessarily approximate, because the internal gating mechanisms in an LSTM each employ a nonlinear activation function, either $\sigma$ or tanh.  \citet{murdoch_beyond_2018} use a linearized approximation $L_{\sigma}$ for $\sigma$ and linearized approximation $L_{\tanh}$ for $\tanh$ such that for arbitrary input $\sum_{j=1}^{N} y_j$: 
\begin{equation}\label{eqn:linear}
\sigma{\left(\sum_{j=1}^{N} y_j\right)} = \sum_{j=1}^{N} L_{\sigma}(y_j)
\end{equation}

These approximations are then used to split each gate into components contributed by the previous hidden state $h^{t-1}$ and by the current input $x^t$, for example the input gate $i^t$:
\begin{eqnarray*} 
i^t &=& \sigma(W_i x^t + V_t h^{t-1} + b_i)\\
&\approx& L_{\sigma}(W_i x^t) + L_{\sigma}(V_t h^{t-1}) + L_{\sigma}(b_i)
\end{eqnarray*}{}

This linear form $L_{\sigma}$ is achieved by computing the Shapley value~\citep{shapley1953value} of its parameter, defined as the average difference resulting from excluding the parameter, over all possible permutations of the input summants. To apply Formula~\ref{eqn:linear} to $\sigma{(y_1 + y_2)}$ for a linear approximation of the isolated effect of the summant $y_1$:
$$
L_{\sigma}(y_1) = \frac{1}{2} [(\sigma(y_1) - \sigma(0)) + (\sigma(y_2 + y_1) - \sigma(y_1)) ]
$$

With this function, we can take a hidden state from the previous timestep, decomposed as $h^{t-1} \approx h^{t-1}_{\beta} + h^{t-1}_{\bar{\beta}; \beta \interaction \bar{\beta}}$ and add $x^t$ to the appropriate component. For example, if $x^t$ is in focus, we count it in the relevant function inputs when computing the input gate:
\begin{eqnarray*}
i^t &=& \sigma(W_i x^t + V_t h^{t-1} + b_i)\\
&\approx& \sigma(W_i x^t + V_t (h^{t-1}_{\beta} + h^{t-1}_{\bar{\beta}; \beta \interaction \bar{\beta}}) + b_i)\\
&\approx& [L_{\sigma}(W_i x^t + V_t h^{t-1}_{\beta}) + L_{\sigma}(b_i)]\\ 
&&+ L_{\sigma}(V_t h^{t-1}_{\bar{\beta}; \beta \interaction \bar{\beta}})\\ 
&=& i^t_{\beta} + i^{t}_{\bar{\beta}; \beta \interaction \bar{\beta}}
\end{eqnarray*}{}

Because the individual contributions of the items in a sequence interact in nonlinear ways, this decomposition is only an approximation and cannot exactly compute the impact of a  specific word or words on the  label predicted.  However, the dynamics of LSTMs are roughly linear in  natural settings, as found by \citet{morcos_insights_2018}, who found close linear projections between the activations at each timestep in a repeating sequence and the activations at the end of the sequence. This observation allows CD to linearize hidden states with low approximation error, but the presence of slight nonlinearity forms the basis for our measure of interdependence later on.

We can use softmax to convert the relevant logits $v_{\beta}$ into a probability distribution as $P(Y \mid x_{\beta}) = \textrm{softmax}(v_{\beta})$. This allows us to analyze the effect of input $x_{\beta}$ on the representation of a later element while controlling for the influence of the rest of the sequence. 

In our analyses, CD yielded an approximation error $\frac{\|(v_{\beta} + v_{\bar{\beta}; \beta \interaction \bar{\beta}})-v\|}{\|v\|}<10^{-5}$ at the logits. However, this measurement misses another source of approximation error: the allocation of credit between $\beta$ and the interactions  $\beta \interaction \bar{\beta}$. Changing the sequence out of focus $\bar{\beta}$ might influence $v_{\beta}$, for example, even though the contribution of the words in focus should be mostly confined to the irrelevant vector component. This approximation error is crucial because the component attributed to $\beta \interaction \bar{\beta}$ is central to our measure of interdependence.

\subsection{Interdependence} \label{sec:interdependence}

We frame compositionality in terms of whether the meanings of a pair of words or word subsets can be treated independently. For example, a ``slice of cake'' can be broken into the individual meanings of ``slice'', ``of'', and ``cake'', but an idiomatic expression such as ``piece of cake'', meaning a simple task, cannot be broken into the individual meanings of ``piece'', ``of'', and ``cake''. The words in the idiom have higher \textbf{interdependence}, or reliance on their interactions to build meaning. Another influence on interdependence should be syntactic relation; if you ``happily eat a slice of cake'', the meaning of ``cake'' does not depend on ``happily'', which modifies ``eat'' and is far on the syntactic tree from ``cake''. We will use the nonlinear interactions in contextual decomposition to analyze the interdependence between words alternately considered in focus.

Generally, CD considers all nonlinear interactions between the relevant and irrelevant sets of words to fall under the irrelevant contribution as $\beta \interaction \bar{\beta}$, although other allocations of interactions have been proposed~\cite{jumelet_analysing_2019}.  A fully flat structure for building meaning could lead to a contextual representation that breaks into a linear sum of each word's meaning, which is the simplifying assumption at the heart of CD. 

Given two interacting sets of words to potentially designate as the $\beta$ in focus, $A,B$ such that $A \cap B = \emptyset$, we use a measure of interdependence to quantify the degree to which $A \cup B$ be broken into their individual meanings. With $v_A$ and $v_B$ denoting the relevant contributions of $A$ and $B$ according to CD, and $v_{A,B}$ as the relevant contribution of $A \cup B$, we compute the magnitude of nonlinear interactions, rescaled to control for the magnitude of the representation: 
$$\textrm{interdependence}(A,B) = \frac{\|v_{A \cup
B}-(v_{A}+v_{B})\|_2}{\|v_{A \cup B}\|_2}$$
This quantity is related to probabilistic independence. We would say that events $X$ and $Y$ are independent if their joint probability $P(X,Y) = P(X)P(Y)$. Likewise, the meanings of $A$ and $B$ can be called independent if $v_{A \cup B}=v_{A}+v_{B}$.

    \begin{figure*}
        \centering
        \begin{subfigure}[b]{0.48\textwidth}
            \centering
            \includegraphics[width=\textwidth]{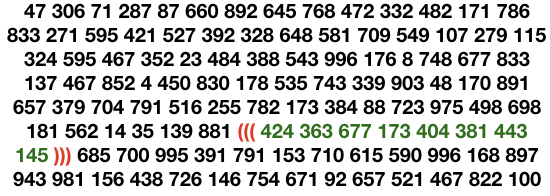}
            \caption[Network2]%
            {{\small unfamiliar-conduit training set}}    
            \label{fig:unpred}
        \end{subfigure}
        \hfill
        \begin{subfigure}[b]{0.48\textwidth}  
            \centering 
            \includegraphics[width=\textwidth]{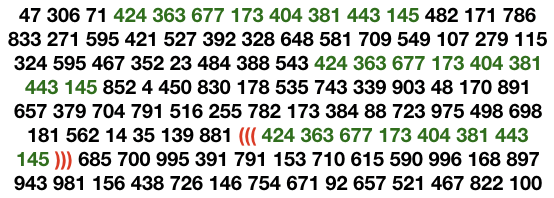}
            \caption[]%
            {{\small familiar-conduit training set}}    
            \label{fig:pred}
        \end{subfigure}
        \vskip\baselineskip
        \begin{subfigure}[b]{0.48\textwidth}   
            \centering 
            \includegraphics[width=\textwidth]{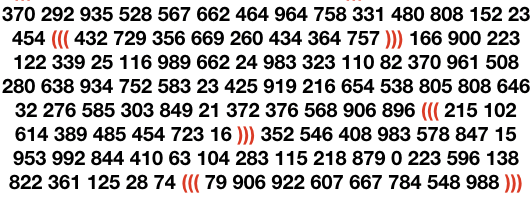}
            \caption[]%
            {{\small out-domain test set}}    
            \label{fig:outdo}
        \end{subfigure}
        \hfill
        \begin{subfigure}[b]{0.48\textwidth}   
            \centering 
            \includegraphics[width=\textwidth]{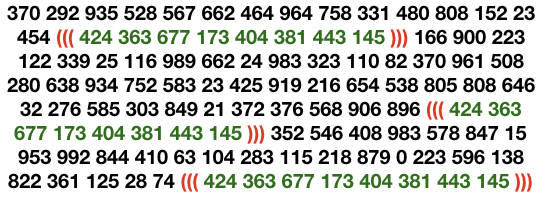}
            \caption[]%
            {{\small in-domain test set}}    
            \label{fig:mean and std of net44}
        \end{subfigure}
        \caption[ The average and standard deviation of critical parameters ]
        {\small We have highlighted rule boundaries {\color{red}$\alpha$} and {\color{red}$\omega$} in red, and conduit {\color{ForestGreen}$q$} $ \in Q_k$ in green.}
        \label{fig:indo}
    \end{figure*}
    
\section{Synthetic Experiments} \label{sec:synthetic}

Our first experiments use synthetic data to understand the role of compositionality in LSTM learning dynamics. These dynamics see long-range connections discovered after short-range connections; in particular document-level content topic information is preserved much later in training than local information like part of speech~\citep{saphra_understanding_2019}. There are several explanations for this phenomenon.

First, long-range connections are less consistent (particularly in a right-branching language like English). For example, the pattern of a determiner followed by a noun will appear very frequently, as in ``the man''. However, we will less frequently see long-range connections like the either/or in ``Either Socrates is mortal or not''. Rarer patterns are learned slowly (Appendix~\ref{sec:frequency}).

The following experiments are designed to explore a third possibility: that the training process is inherently compositional. That is, the shorter sequences must be learned first in order to form the basis for longer relations learned around them. The compositional view of training is not a given and must be verified. In fact, simple rules learned early on might inhibit the learning of more complex rules through the phenomenon of gradient starvation \cite{combes_learning_2018}, in which more frequent features dominate the gradient directed at rarer features.  Shorter familiar patterns could slow down the process for learning longer range patterns by degrading the gradient passed through them, or by trapping the model in a local minimum which makes the long-distance rule harder to reach. However, if the training process builds syntactic patterns hierarchically, it can lead to representations that are built hierarchically at inference time, reflecting linguistic structure. To test the idea of a compositional training process, we use synthetic data that controls for the consistency and frequency of longer-range relations.




\subsection{The dataset}

\begin{figure*}
        \centering
        \begin{subfigure}[b]{0.48\textwidth}
            \centering
            \includegraphics[width=\textwidth]{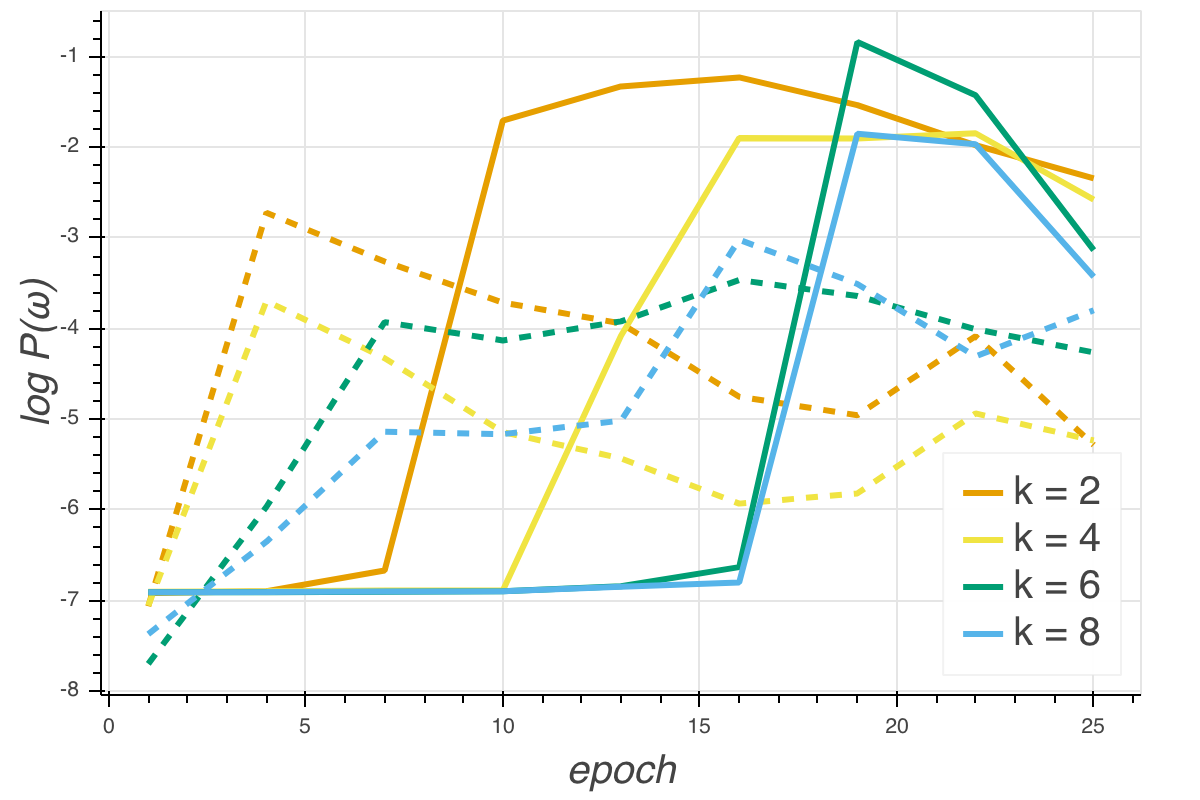}
            \caption[Network2]%
            {{\small In-domain conduit test setting}}    
            \label{fig:indomain}
        \end{subfigure}
        \hfill
        \begin{subfigure}[b]{0.48\textwidth}  
            \centering 
            \includegraphics[width=\textwidth]{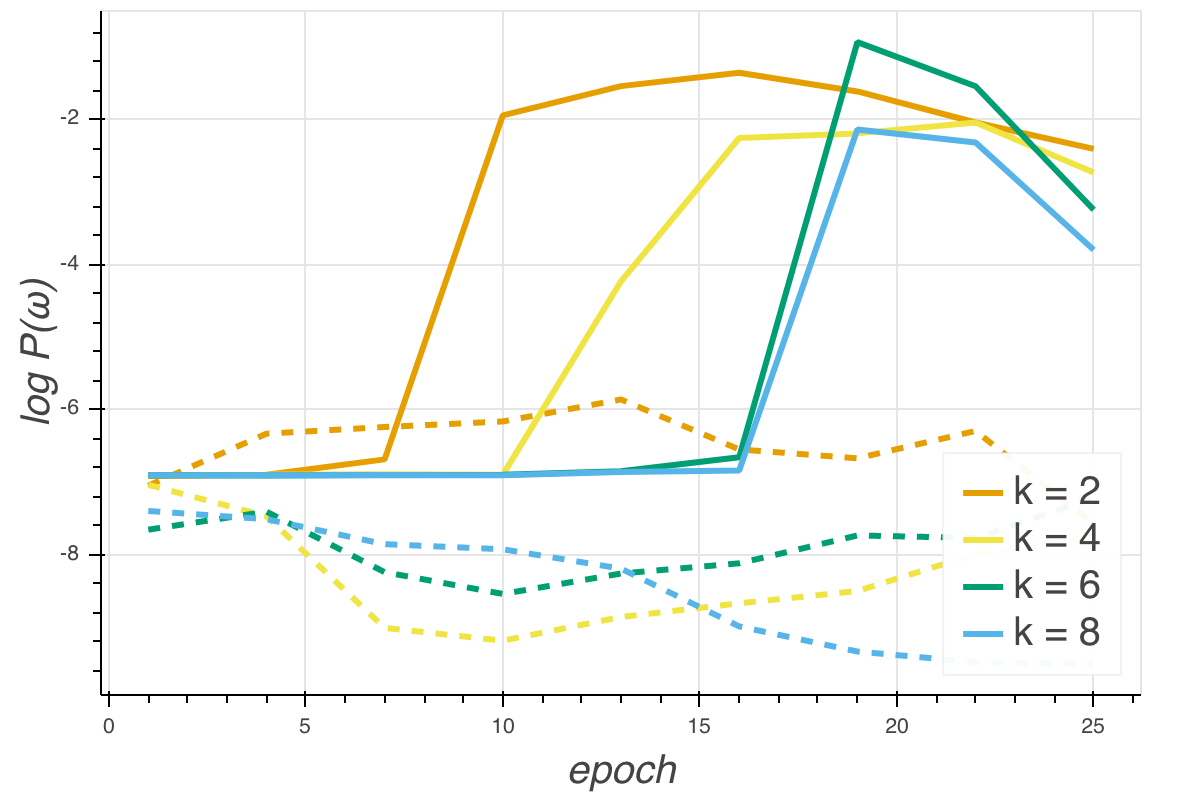}
            \caption[]%
            {{\small Random conduit test setting}}    
            \label{fig:outdomain}
        \end{subfigure}
\caption{Mean marginal target probability of the close symbol in a rule. Solid lines are trained in the unfamiliar-conduit set, dashed lines on familiar-conduit. Scale of y-axis is matched among graphs.}
\end{figure*}
We may expect representations of long-range connections to be built in a way that depends strongly on the subtrees they span, but not all such connections rely on subtrees. If connections that do not rely on shorter constituents nonetheless are built from these constituents, then the observation over training, that increasingly long connections are gradually constructed from constituents, cannot be dismissed as purely a data effect. 
For example, consider ``either/or''. ``Either'' should determine that ``or'' will later occur, regardless of the phrase that intercedes them. To learn this rule, a language model must backpropagate information from the occurrence of ``or'' through the intervening sequence of words, which we will call a \textbf{conduit}. Perhaps it encounters a training example that uses  a conduit that is predictable by being structured in familiar ways,  here italicized: ``Either  \textit{Socrates is  mortal} or not''. But what if the conduit is unfamiliar and the structure cannot be interpreted by the model? For example, if the conduit includes unknown tokens: ``Either \textit{slithy toves gyre} or not''. How will the gradient carried from ``or'' to ``either'' be shaped according to the conduit, and how will the representation of that long-range connection change accordingly? A familiar conduit could be used by a compositional training process as a short constituent on which to build longer-range representations, so the meaning of ``Either'' in context will depend on the conduit. Conversely, if training is not biased to be compositional, the connection will be made regardless so the rule will generalize to test data. To investigate whether long-range dependencies are built from short constituents in this way, we train models on synthetic data which varies the predictability of short sequences.

We generate data uniformly at random from  a vocabulary $\Sigma$. We insert $n$ instances of the long-distance rule $\alpha \Sigma^k \omega$, with conduit $\Sigma^k$ of length $k$, \textbf{open symbol} $\alpha$, and \textbf{close symbol} $\omega$, with $\alpha, \omega \not\in \Sigma$. Relating to our running example, $\alpha$ stands for ``either'' and $\omega$ stands for ``or''. We use a corpus of 1m tokens with $|\Sigma|=$ 1k types, which  leaves a low probability that any conduit sequence longer than 1 token appears elsewhere by chance.

We train with a learning rate set at 1 throughout and gradients clipped at 0.25. We found momentum and weight decay to slow rule learning in this setting, so they are not used.

\subsection{The Effect of Conduit Familiarity}

To understand the effect of conduit predictability on longer-range connections, we modify the original synthetic data (Figure~\ref{fig:unpred}) so each conduit appears frequently outside of the $\alpha/\omega$ rule (Figure~\ref{fig:pred}). The conduits are sampled from a randomly generated vocabulary of 100 phrases of length $k$, so each unique conduit $q$ appears in the training set 10 times in the context $\alpha q\omega$. This repetition is necessary in order to fit $1000$ occurrences of the rule in all settings. In the \textbf{familiar-conduit setting}, we randomly distribute $1000$ occurrences of each conduit throughout the corpus outside of the rule patterns. Therefore each conduit is seen often enough to be memorized (see Appendix~\ref{sec:bptt}). In the original \textbf{unfamiliar-conduit setting}, $q$ appears only in this context as a conduit, so the conduit is not memorized.

We also use two distinct test sets. Our in-domain test set (Figure~\ref{fig:indo}) uses the same set of conduits as the train set. In Figure~\ref{fig:indomain}, the model learns when to predict the close symbol faster if the conduits are familiar (as predicted in Appendix~\ref{sec:bptt}). However, the maximum performance reached is lower than with the unfamiliar setting, and is brittle to overtraining.

If the test set conduits are sampled uniformly at random (Figure~\ref{fig:outdo}), Figure~\ref{fig:outdomain} shows that the familiar-conduit training setting never teaches the model to generalize the $\alpha/\omega$ rule. For a model trained on the familiar domain, a familiar conduit is required to predict the close symbol.

\begin{figure}
    \centering
    \includegraphics[width=0.48\textwidth]{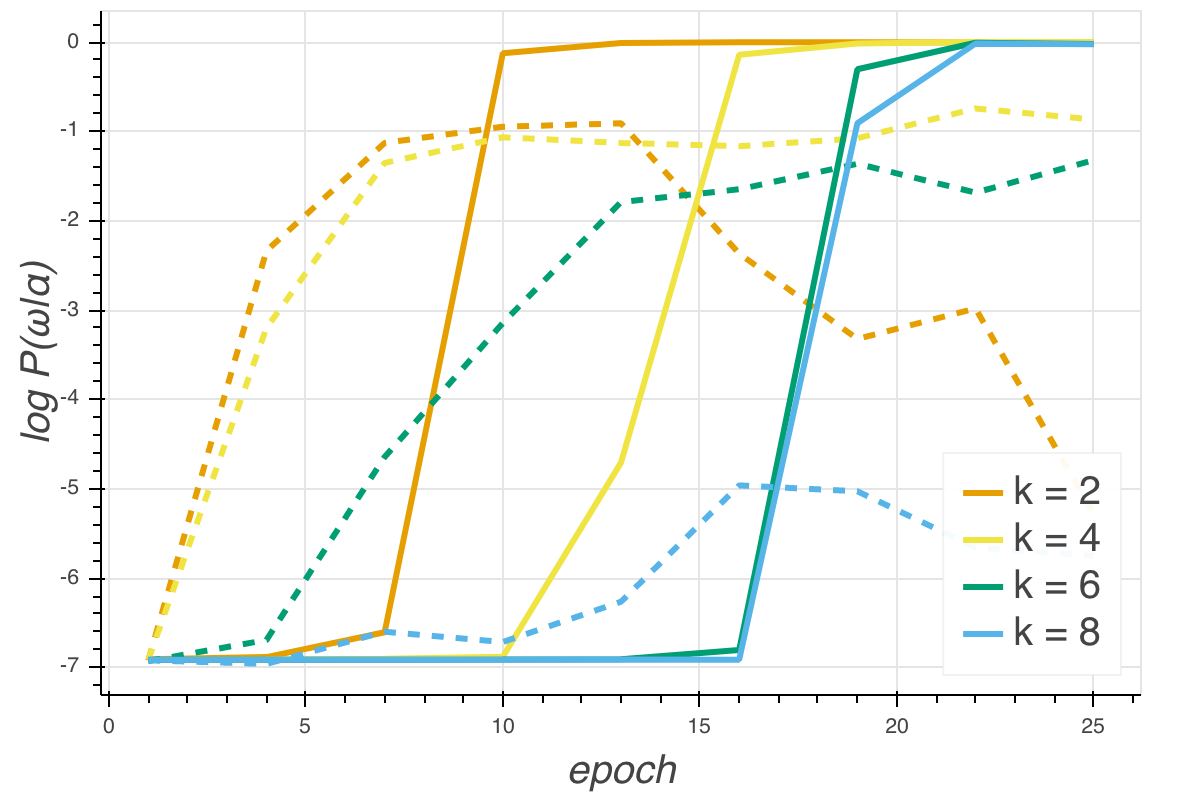}
    \caption{Mean target probability of $\omega$ using CD with $\alpha$ in focus, out-domain test set. Solid lines are trained in the unfamiliar-conduit set, dashed lines on familiar-conduit.}
    \label{fig:cd_alpha}
\end{figure}

\subsubsection{Isolating the Effect of the Open-Symbol}

Raw predictions in the out-domain test setting appear to suggest that the familiar-conduit training setting fails to teach the model to associate $\alpha$ and $\omega$. However, the changing domain makes this an unfair assertion: the poor performance may be attributed to interactions between the open symbol and the conduit. In order to control for the potential role that memorization of conduits plays in the  prediction of the close symbol, we use CD to isolate  the  contributions of the open symbol in the random conduit test setting.

\begin{figure}
    \centering
    \includegraphics[width=0.48\textwidth]{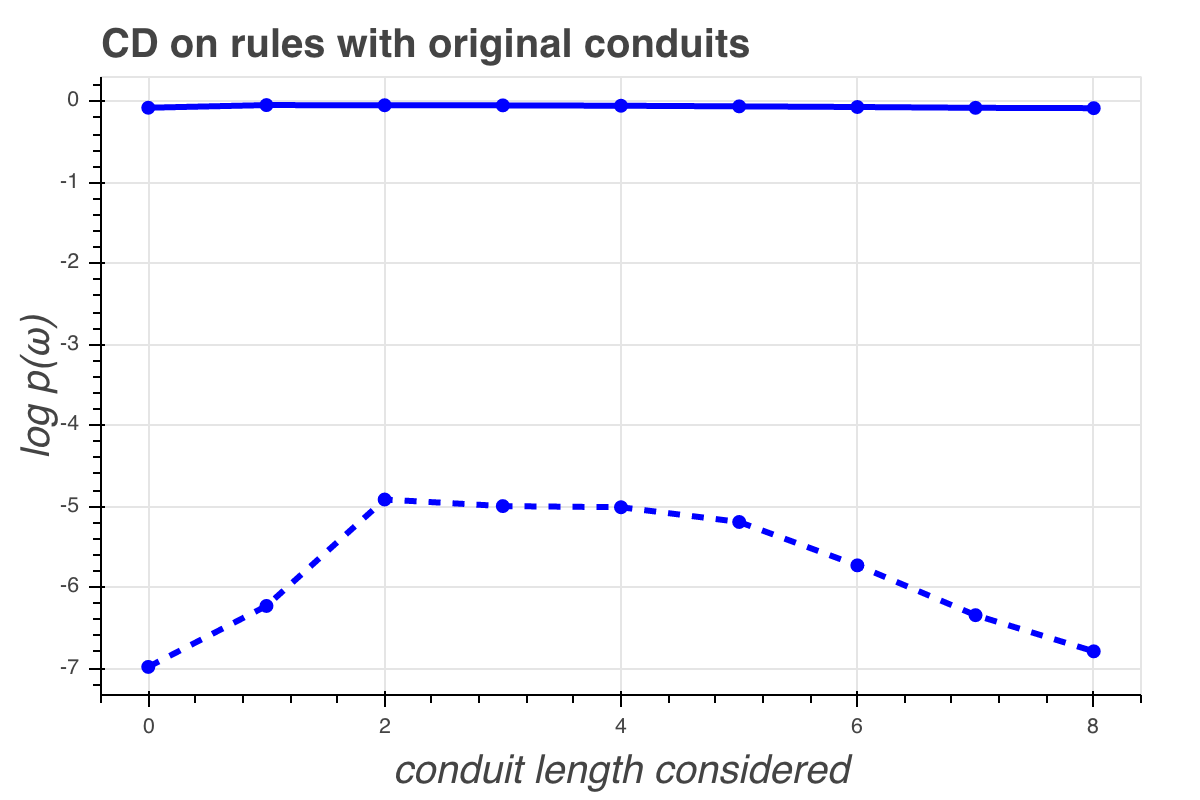}
    \caption{The predicted $P(x_{t} = \omega | x_{t-k} \ldots x_{t-k+i})$ according to CD, varying $i$ as the x-axis and with $x_{t-k} = \alpha$ and $k=8$. Solid lines are trained in the unfamiliar-conduit set, dashed lines on familiar-conduit.}
    \label{fig:incremental}
\end{figure}

Figure~\ref{fig:cd_alpha} shows that even in the out-domain test setting, the presence of $\alpha$ predicts $\omega$ at the appropriate time step. Furthermore, we confirm that the familiar-conduit training setting enables earlier acquisition of this rule.

To what, then, can we attribute the failure to generalize to the random-conduit domain? Figure~\ref{fig:incremental} illustrates how the unfamiliar-conduit model predicts the close symbol $\omega$ with high probability based only on the contributions of the open symbol $\alpha$. Meanwhile, the familiar-conduit model probability increases substantially with each symbol consumed until the end of the conduit, indicating that the model is relying on interactions between the open symbol and the conduit rather than registering only the effect of the open symbol. Note that this effect cannot be because the conduit is more predictive of $\omega$. Because each conduit appears frequently outside of the specific context of the rule in the familiar-conduit setting, the conduit is \textit{less} predictive of $\omega$ based on distribution alone.

\begin{figure}
    \centering
    \includegraphics[width=0.48\textwidth]{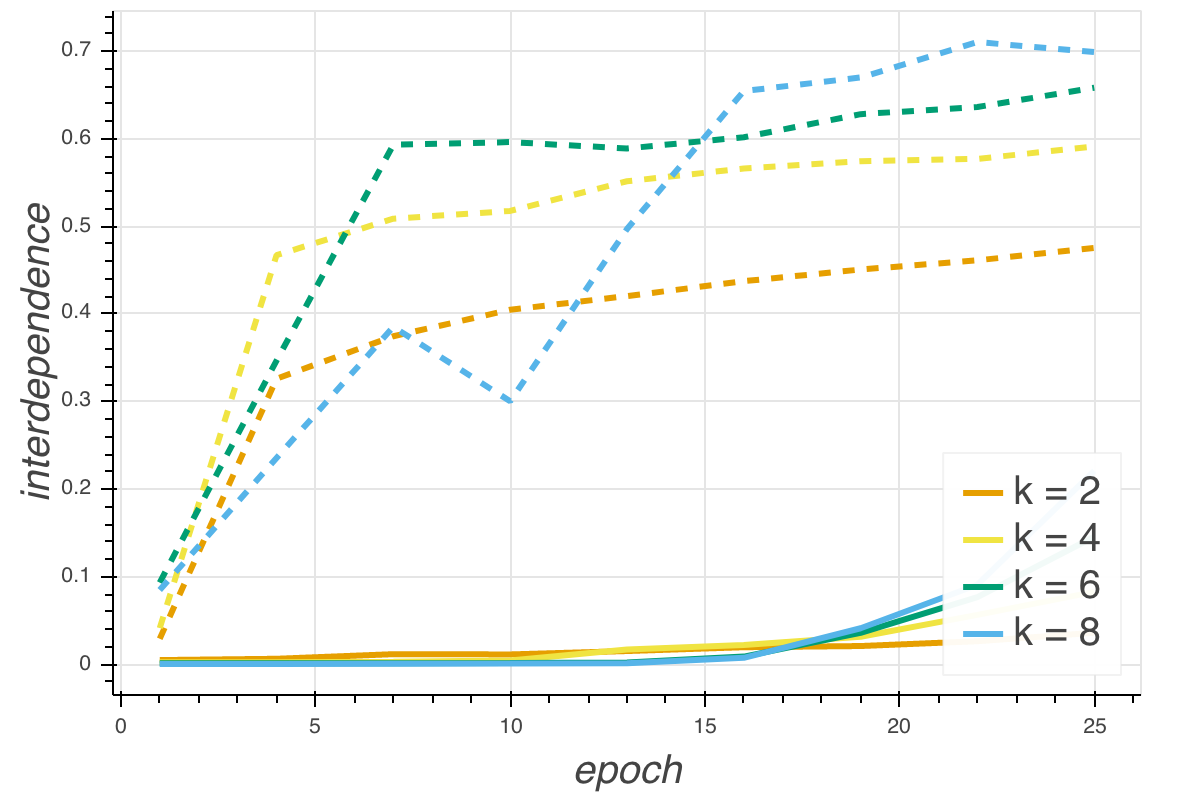}
    \caption{Mean interdependence between open symbol and conduit on the in-domain test set. Solid lines are trained in the unfamiliar-conduit set, dashed lines on familiar-conduit.}
    \label{fig:interdependence}
\end{figure}

These results indicate that predictable patterns play a vital role in shaping the representations of symbols around them by composing in a way that cannot be easily linearized as a sum of the component parts. In particular, as seen in Figure~\ref{fig:interdependence}, the interdependence between open symbol and conduit is substantially higher for the familiar-setting model and increases throughout training. Long-range connections are not learned independently from conduit representations, but are \textit{built compositionally} using already-familiar shorter subsequences as scaffolding.

{}

\section{English Language Experiments} \label{sec:english}

\begin{figure}
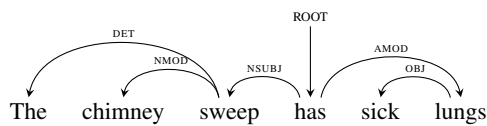
\begin{center}
    \begin{dependency}[theme = simple]\small
        \begin{deptext}[column sep=1em]
          The \& chimney \& sweep \& has \& sick \& lungs \\
        \end{deptext}
    \deproot[edge unit distance=2ex]{4}{ROOT}
    \depedge{3}{1}{\textsc{det}}
    \depedge{3}{2}{\textsc{nmod}}
    \depedge{4}{3}{\textsc{nsubj}}
    \depedge{4}{6}{\textsc{amod}}
    \depedge{6}{5}{\textsc{obj}}
    \end{dependency}\end{center}
    \caption{A dependency parsed sentence.} \label{fig:chimney}
\end{figure}

We now apply our measure of interdependence to a natural language setting. In natural language, disentangling the meaning of individual words requires contextual information which is hierarchically composed. For example, in the sentence, ``The chimney sweep has sick lungs'', ``chimney sweep'' has a clear definition and strong connotations that are less evident in each word individually. However, knowing that ``sweep'' and ``sick'' co-occur is not sufficient to clarify the meaning and connotations of either word or compose a shared meaning. Does interdependence effectively express this syntactic link?

These experiments use language models trained on wikitext-2~\cite{merity2016pointer}, run on the Universal Dependencies corpus English-EWT~\cite{silveira14gold}.

\subsection{Interdependence and Syntax}

\begin{figure}
    \centering
    \includegraphics[width=0.48\textwidth]{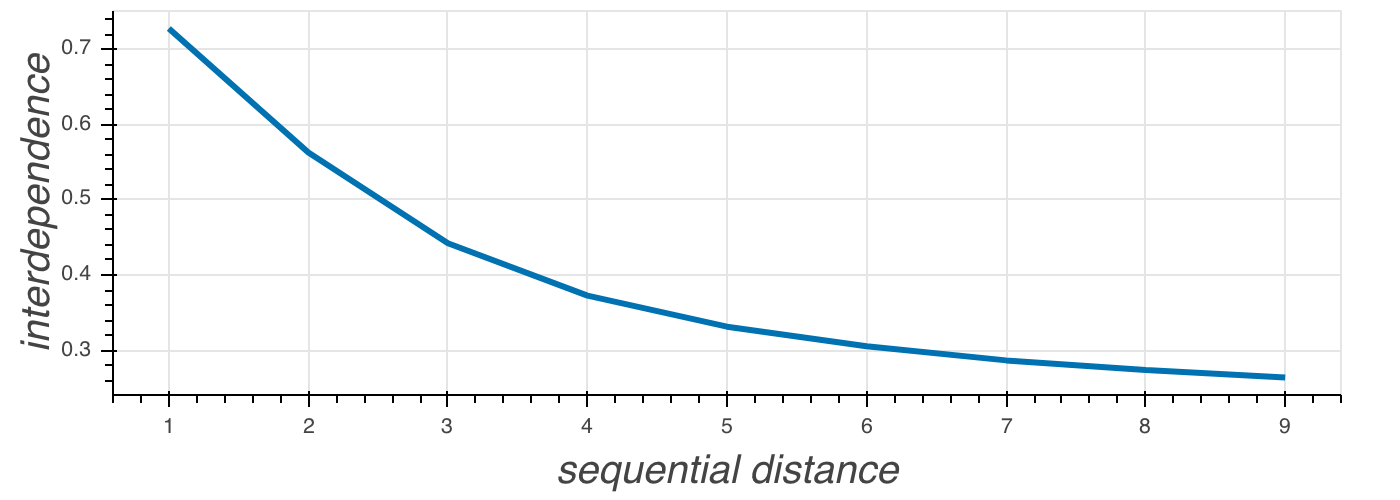}
    \caption{Average interdependence between word pairs $x_l, x_r$ at different sequential distances $r-l$.}
    \label{fig:interdep_all}
\end{figure}

To assess the connection between interdependence and syntax, we consider the interdependence of word pairs with different syntactic distances. For example, in Figure~\ref{fig:chimney}, ``chimney'' is one edge away from ``sweep'', two from ``has'', and four from ``sick''. In Figure~\ref{fig:interdep_all}, we see that in general, the closer two words occur in sequence, the more they influence each other, leading to correspondingly high interdependence. Because proximity is a dominant factor in interdependence, we control for the sequential distance of words when we investigate syntactic distance.

\begin{figure*}
        \centering
        \begin{subfigure}[b]{0.18\textwidth}
            \centering
            \includegraphics[width=\textwidth]{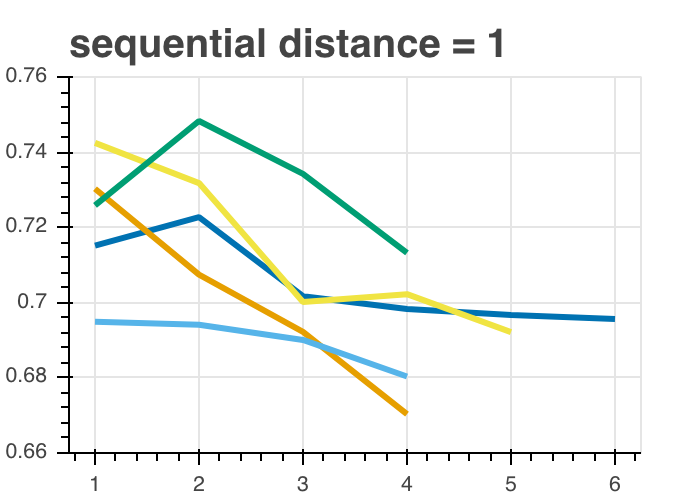}
            
        \end{subfigure}
        \hfill
        \begin{subfigure}[b]{0.18\textwidth}
            \centering
            \includegraphics[width=\textwidth]{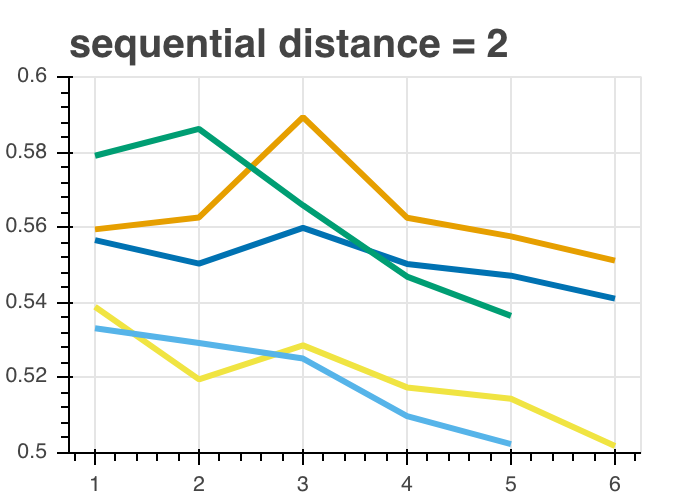}
            
        \end{subfigure}
        \hfill
        \begin{subfigure}[b]{0.18\textwidth}
            \centering
            \includegraphics[width=\textwidth]{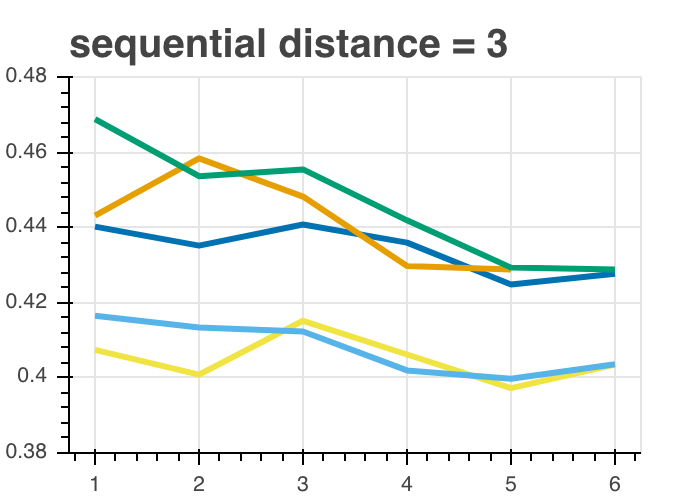}
            
        \end{subfigure}
        \hfill
        \begin{subfigure}[b]{0.18\textwidth}
            \centering
            \includegraphics[width=\textwidth]{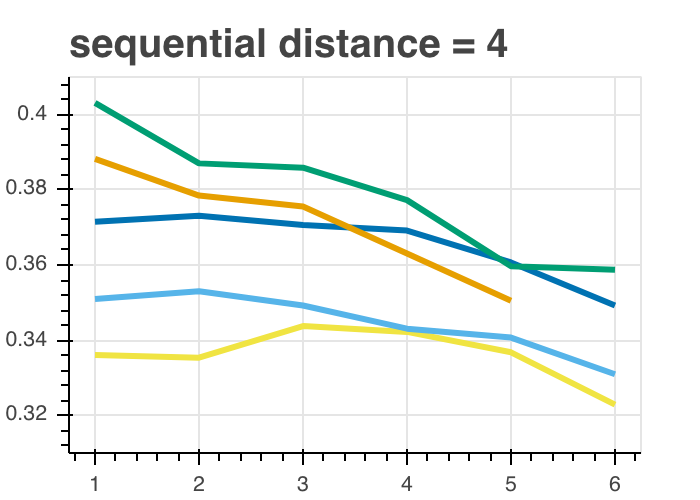}
            
        \end{subfigure}
        \hfill
        \begin{subfigure}[b]{0.18\textwidth}  
            \centering 
            \includegraphics[width=\textwidth]{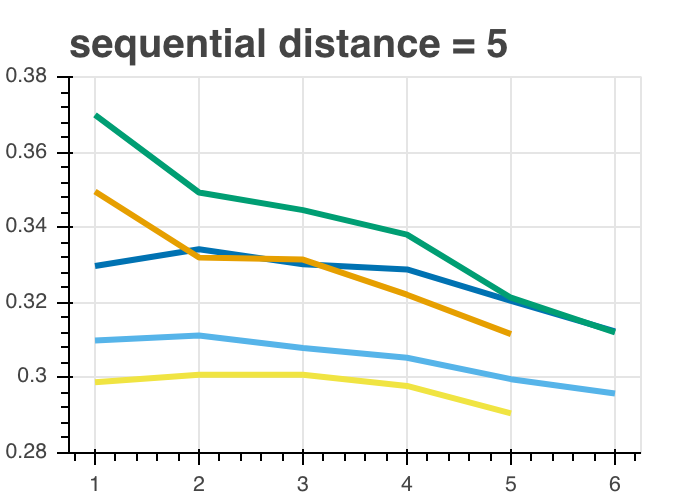}
            
        \end{subfigure}
        \vskip\baselineskip
        \begin{subfigure}[b]{0.18\textwidth}
            \centering
            \includegraphics[width=\textwidth]{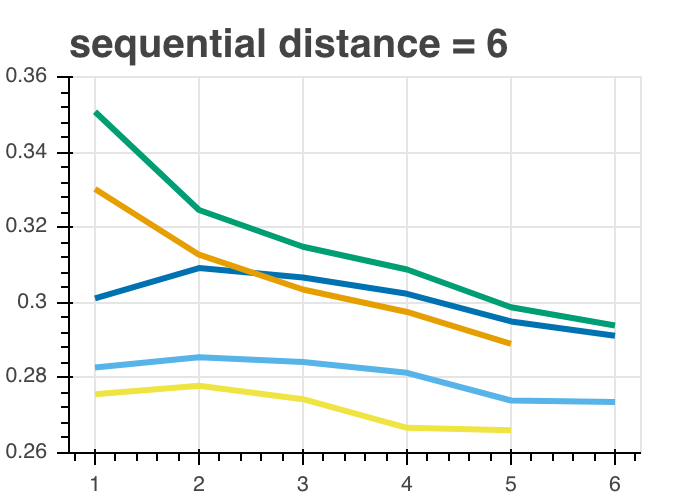}
        \end{subfigure}
        \hfill
        \begin{subfigure}[b]{0.18\textwidth}
            \centering
            \includegraphics[width=\textwidth]{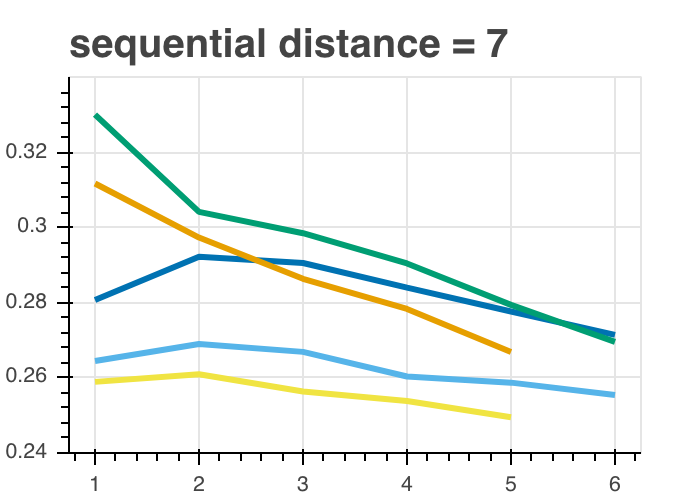}
        \end{subfigure}
        \hfill
        \begin{subfigure}[b]{0.18\textwidth}
            \centering
            \includegraphics[width=\textwidth]{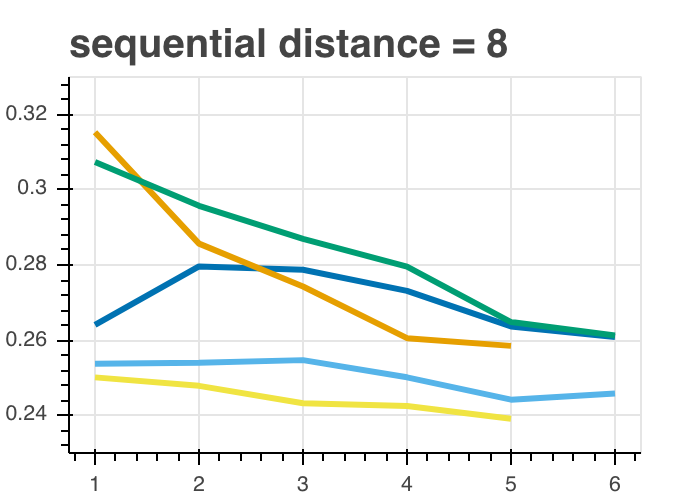}
        \end{subfigure}
        \hfill
        \begin{subfigure}[b]{0.18\textwidth}
            \centering
            \includegraphics[width=\textwidth]{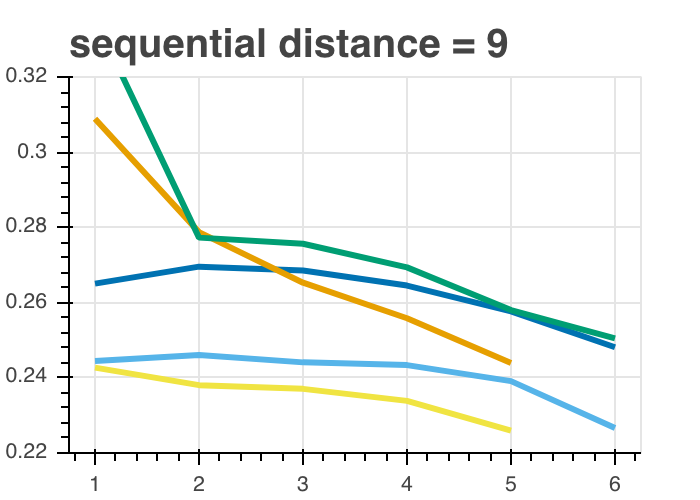}
        \end{subfigure}
        \hfill
        \begin{subfigure}[b]{0.18\textwidth}  
            \centering 
            \includegraphics[width=\textwidth]{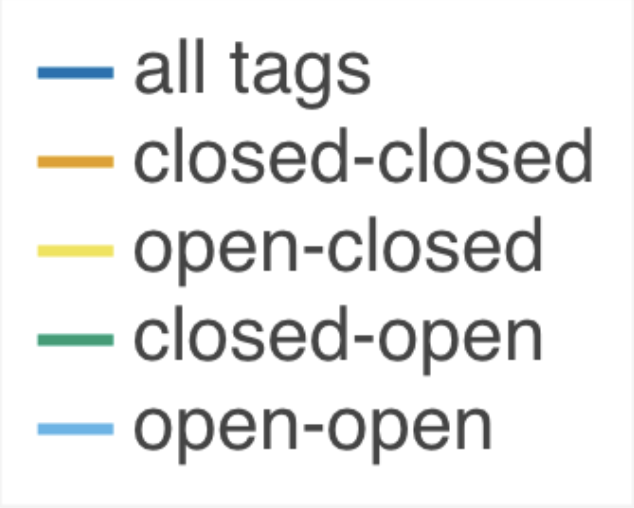}
        \end{subfigure}
\caption{Mean interdependence (y-axis) between word pairs at varying syntactic distances (x-axis), stratified by whether the POS tags are closed or open class (line color) and by sequential distance (plot title). The y-axis ranges differ, but the scale is the same for all plots. Each mean is plotted only if there are at least 100 cases to average.}
\label{fig:closed_and_open_tags}
\end{figure*}

The synthetic data experiments show that phrase frequency and predictability play a critical role in determining interdependence (although raw word frequency shows no clear correlation with interdependence in English). We control for these properties through POS tag, as open and closed tags vary in their predictability in context; for example, determiners (a closed POS class) are almost always soon followed by a noun, but adjectives (an open POS class) appear in many constructions like ``Socrates is mortal'' where they are not. We stratify the data in Figure~\ref{fig:closed_and_open_tags} according to whether the POS tags are in closed or open classes, which serves as a proxy for predictability. Irrespective of both sequential distance and part of speech, we see broadly decreasing trends in interdependence as the syntactic distance between words increases, consistent with the prediction that syntactic proximity drives interdependence. This pattern is clearer as words become further apart in the sequence, implicating non-syntactic influences such as priming effects that are stronger immediately following a word.

{}

\section{Discussion \& Related Work}

Humans learn by memorizing short rote phrases and later mastering   the ability to construct deep syntactic trees from them \citep{lieven2008children}. LSTM models learn by backpropagation through time, which is unlikely to lead to the same \textit{inductive biases}, the assumptions that define how the model generalizes from its training data. It may not be expected for an LSTM to exhibit similarly compositional learning behavior by building longer constituents out of shorter ones during training, but we present evidence in favor of such learning dynamics.

LSTMs have the theoretical capacity to encode a wide range of context-sensitive languages, but in practice their ability to learn such rules from data is limited \citep{weiss_practical_2018}. Empirically, LSTMs encode the most recent noun as the subject of a verb by default,  but they are still capable of learning to encode grammatical inflection from the first word in a sequence rather than the most recent \cite{ravfogel_studying_2019}. Therefore, while inductive biases inherent to the model play a critical role in the ability of an LSTM to learn effectively, they are neither necessary nor sufficient in determining what the model can learn. Hierarchical linguistic structure may be learned from data alone, or be a natural product of the training process, with neither hypothesis a foregone conclusion. We provide a more precise lens on how LSTM training is itself compositional.

While \citet{saphra_understanding_2019} illustrate LSTM learning dynamics expanding from representing short-range properties to long, \citet{voita_bottom-up_2019} presents evidence of transformer models building long range connections after shorter ones. However, it is not clear whether this compositionality is an inherent property of the model or an effect of hierarchical structure in the data.

There is a limited literature on compositionality as an inductive bias of neural networks. 
\citet{saxe_mathematical_2018} explored how hierarchical ontologies are learned by following their tree structure in 2-layer feedforward networks. \citet{liu_lstms_2018} showed that LSTMs take advantage of some inherent trait of language. The compositional training we have explored may be the mechanism behind this biased representational power.

Synthetic data, meanwhile, has formed the basis for analyzing the inductive biases of neural networks their capacity to learn compositional rules. Common synthetic datasets include the Dyck languages~\citep{suzgun-etal-2019-lstm,skachkova-etal-2018-closing}, SPk~\citep{mahalunkar-kelleher-2019-multi}, synthetic variants of natural language~\citep{DBLP:journals/corr/abs-1903-06400,liu_lstms_2018}, and others~\citep{DBLP:journals/corr/abs-1906-00180,livska2018memorize,korrel-etal-2019-transcoding}. Unlike these works, our synthetic task is not designed primarily to test the biases of the neural network or to improve its performance in a restricted setting, but to investigate the internal behavior of an LSTM in response to memorization.

Our results may offer insight into selecting training curricula. For feedforward language models, a curriculum-based approach has been shown to improve performance  from small training data~\cite{bengio_curriculum_2009}. Because LSTMs seem to naturally learn short-range dependencies before long-range dependencies, it may be tempting to enhance this natural tendency with a curriculum. However, curricula that move from short sequences to long apparently fail to support more modern recurrent language models. Although these data schedules may help the model converge faster and improve performance early on \cite{zhang_boosting_2017}, after further training the model underperforms against shuffled baselines \cite{zhang_empirical_2018}. Why? We propose the following explanation.

The application of a curriculum is based on the often unspoken assumption that the representation of a complex pattern can be reached more easily from a simpler pattern. However, we find that effectively representing shorter conduits actually makes a language model \textit{less} effective at generalizing a long-range rule. However, this less generalizable representation is still learned faster, which may be why \citet{zhang_boosting_2017} found higher performance after one epoch. Our work suggests that measures of length, including syntactic depth, may be inappropriate bases for curriculum learning.

\section{Future Work}

While we hope to isolate the role of long range dependencies through synthetic data, we must consider the possibility that the natural predictability of language data differs in relevant ways from the synthetic data, in which the conduits are predictable only through pure memorization. Because LSTM models take advantage of linguistic structure, we cannot be confident that predictable natural language exhibits the same cell state dynamics that make a memorized uniformly sampled conduit promote or inhibit long-range rule learning. Future work could test these findings through carefully selected natural language, rather than synthetic, data. 

Our natural language results could lead to interdependence as a probe for testing syntax. Similar analyses of word interaction and nonlinearity may also be used to probe transformer models. 

Some effects on our natural language experiments may be due to the predictable nature of English syntax, which favors right-branching behavior. Future work could apply similar analysis to other languages with different grammatical word orders.

\section{Conclusions}


With synthetic experiments, we confirm that the longer the span of a rule, the more examples are required for an LSTM model to effectively learn the rule. We then find that a more predictable conduit between rule symbols promotes early learning of the rule, but fails to generalize to new domains, implying that these memorized patterns lead to representations that depend heavily on interactions with the conduit instead of learning the long-distance rule in isolation. We develop a measure of interdependence to quantify this reliance on interactions. In natural language experiments, we find higher interdependence indicates words that are closer in the syntax tree, even stratified by sequential distance and part of speech.

\bibliography{acl2019}
\bibliographystyle{acl_natbib}

\clearpage
\newpage
\appendix

\section{The Effect of Rule Frequency and Length} \label{sec:frequency}
\begin{figure*}
    \centering
    \includegraphics[width=0.32\textwidth]{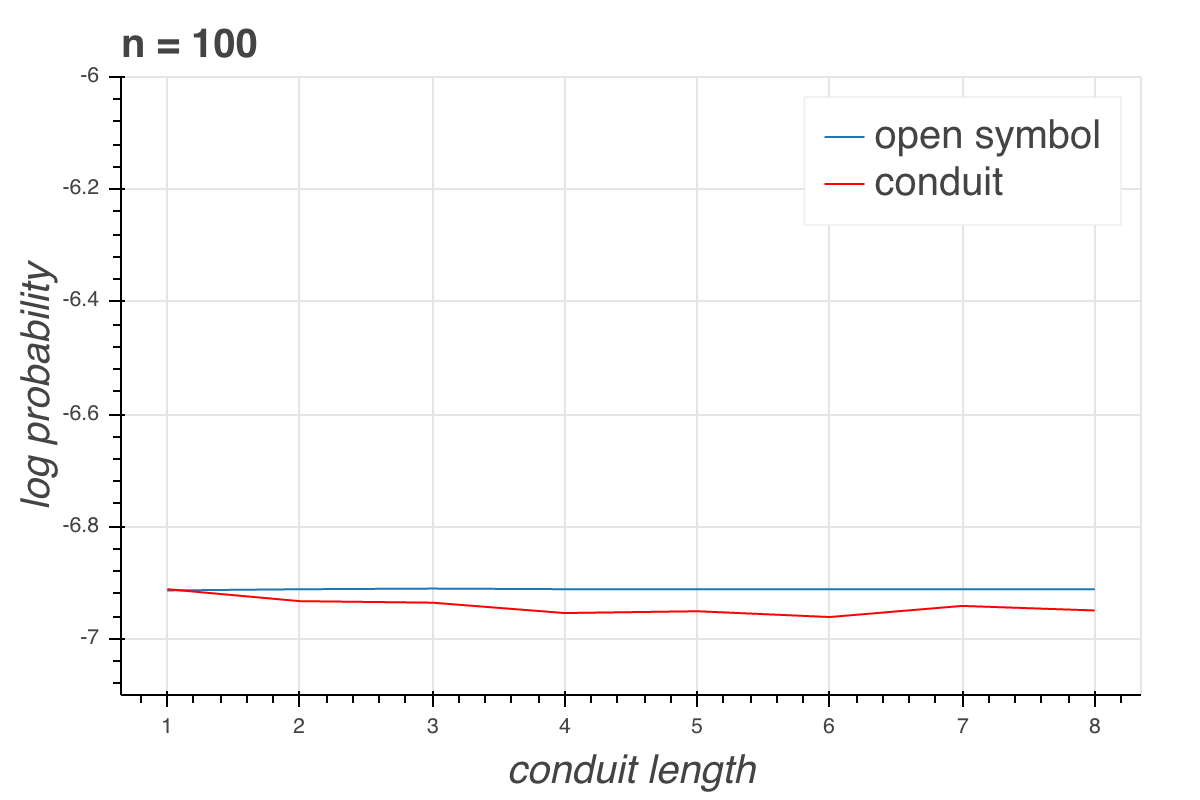}
    \includegraphics[width=0.32\textwidth]{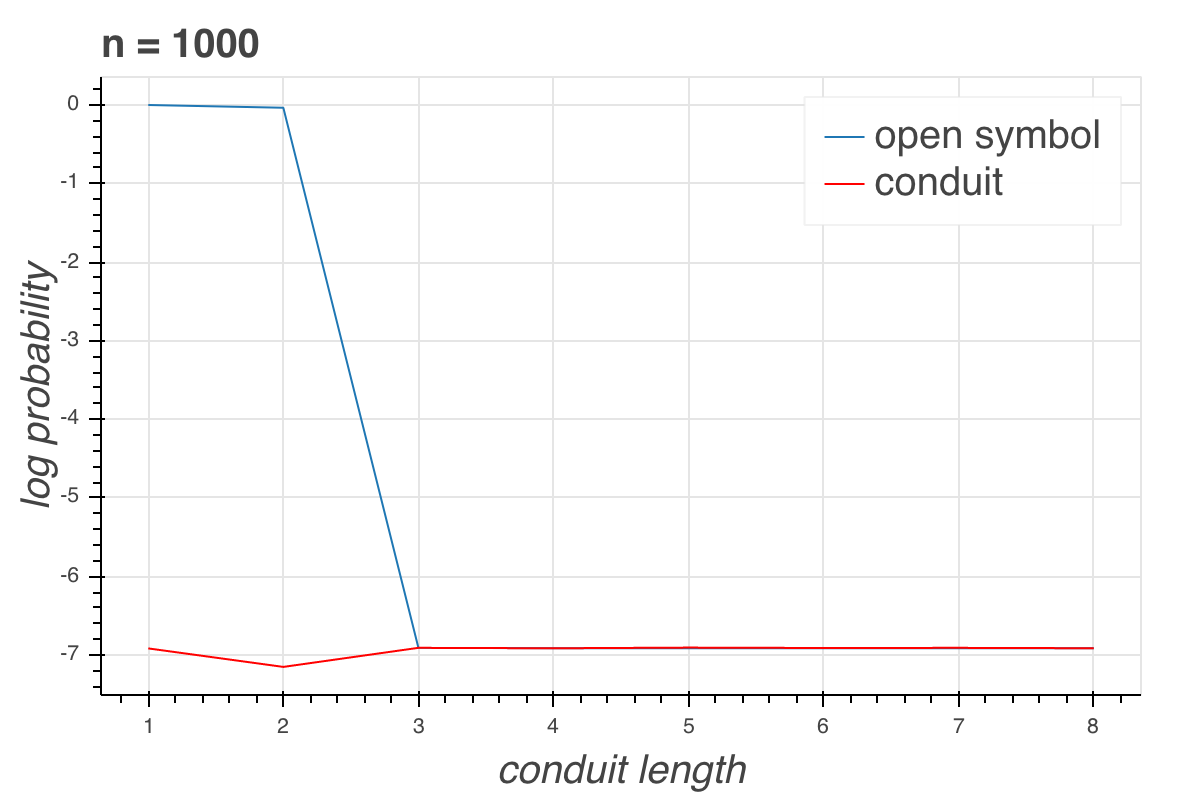}
    \includegraphics[width=0.32\textwidth]{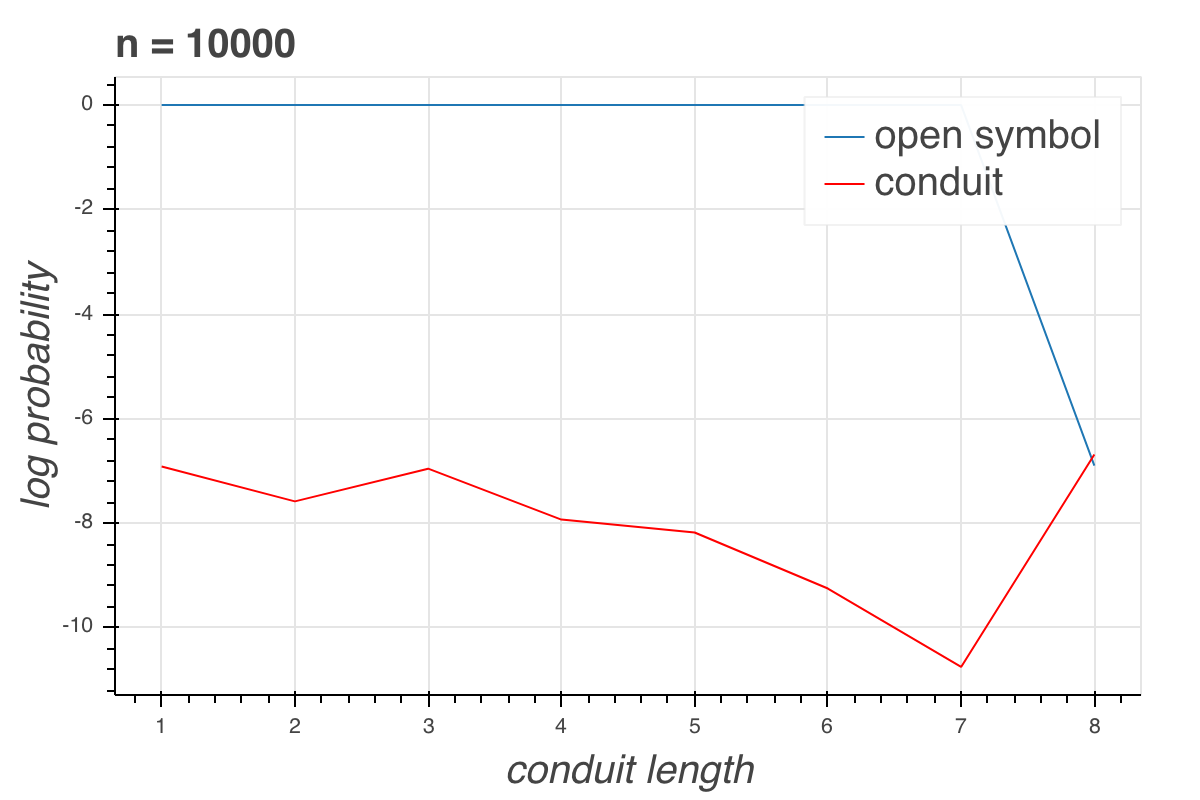}
    \caption{The predicted probability $P(x_{t} = \omega)$, according to the contributions of open symbol $x_{t-k} = \alpha$ and of the conduit sequence $x_{t-k+1} \ldots x_{t-1}$, for various rule occurrence counts $n$. Shown at 40 epochs.}
    \label{fig:rule_frequency}
\end{figure*}

 We investigate how the frequency of a rule affects the ability of the model to  learn the rule. We vary the number of rule occurrences $n$ and the rule length $k$. The results in Figure~\ref{fig:rule_frequency} illustrate how a longer conduit length requires more  examples before the model can learn the corresponding  rule.  We consider the probability assigned to the close symbol according to the contributions of the open symbol, excluding interaction from any other token in the sequence. For contrast, we also show the extremely low probability assigned to the close symbol according to the contributions of the conduit taken as an entire phrase. In particular, note the pattern when the rule is extremely rare:  The probability of the close symbol  as determined by the open symbol is low but steady, while the probability  as determined by the conduit declines with conduit length  due to the accumulated low probabilities from each element in the sequence.

\section{Smaller conduit gradient, faster rule learning} \label{sec:bptt}

\begin{figure}
\includegraphics[width=\linewidth]{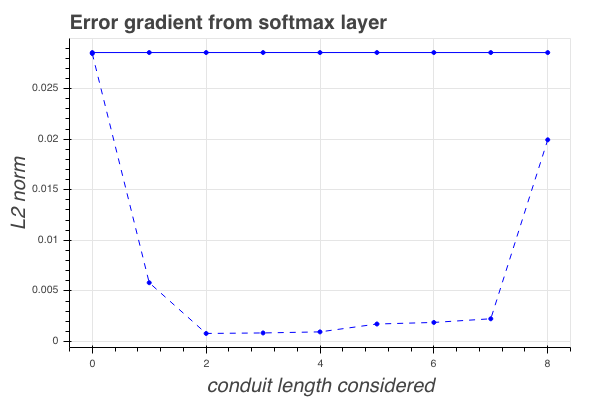}
\caption{Average gradient magnitude $\Delta E_{t+-k+d}$, varying $d$ up to the length of the conduit. Solid lines are the unpredictable conduit setting, dashed lines are the predictable conduit setting.}
\label{fig:error}
\end{figure}

Figure~\ref{fig:error} confirms that a predictable conduit is associated with a smaller error gradient. Because of the mechanics of backpropagation through time next described, this setting will teach the $\alpha/\omega$  rule faster. 

Formally in a simple RNN, as the gradient of the error $e_t$ at timestep $t$ is backpropagated $k$ timesteps through the hidden state $h$:
$$\frac{\partial e_t}{\partial h_{t-k}} = \frac{\partial e_t}{\partial h_t} \prod_{i=1}^k \frac{\partial h_{t-i+1}}{\partial h_{t-i}}$$ 
The backpropagated message is multiplied repeatedly by the gradient at each timestep in the conduit. If the recurrence derivatives $\frac{\partial h_{i+1}}{\partial h_{i}}$ are large at some weight, the correspondingly larger backpropagated gradient $\frac{\partial e_t}{\partial h_{t-k}}$ will accelerate descent at that parameter. 
In other words, an unpredictable conduit associated with a high error will dominate the gradient's sum over recurrences, delaying the acquisition of the symbol-matching rule. In the case of an LSTM, \citet{kanuparthi_h-detach:_2018} expressed the backpropagated gradient as an iterated addition of the error from each timestep, leading to a similar effect. 
\end{document}